\documentclass[letterpaper,10pt]{article}

\usepackage[accepted]{eiml_icml2026}
\usepackage{amsmath,amssymb,amsthm,thmtools,mathtools}
\declaretheorem[name=Assumption,numberwithin=section]{assumption}
\declaretheorem[name=Proposition,numberwithin=section]{proposition}
\usepackage{bm,bbm}
\usepackage{booktabs}
\usepackage{caption}
\usepackage{hyperref}
\usepackage{cleveref}
\crefname{assumption}{Assumption}{Assumptions}
\crefname{theorem}{Theorem}{Theorems}
\crefname{proposition}{Proposition}{Propositions}
\crefname{lemma}{Lemma}{Lemmas}
\crefname{corollary}{Corollary}{Corollaries}
\usepackage{algorithm,algorithmic}
\usepackage{graphicx,subcaption}
\usepackage{xcolor,microtype}
\usepackage{enumitem}
\usepackage{stfloats}
\usepackage{multirow}
\newcommand{\bh}{\bm{h}}
\newcommand{\bw}{\bm{w}}
\newcommand{\bmu}{\bm{\mu}}
\newcommand{\bSigma}{\bm{\Sigma}} 
\newcommand{\bH}{\bm{H}}
\newcommand{\bI}{\bm{I}}
\newcommand{\by}{\bm{y}}
\newcommand{\R}{\mathbb{R}}
\newcommand{\E}{\mathbb{E}}

\newcommand{\Dtrain}{\mathcal{D}_{\mathrm{train}}}
\newcommand{\Dcal}{\mathcal{D}_{\mathrm{cal}}}
\newcommand{\Dweight}{\mathcal{D}_{\mathrm{weight}}}
\newcommand{\Dtest}{\mathcal{D}_{\mathrm{test}}}
\newcommand{\Ncal}{\mathcal{N}}
\newcommand{\ps}{p_{\mathrm{s}}}
\newcommand{\pt}{p_{\mathrm{t}}}
\newcommand{\acp}{\alpha_{\mathrm{cp}}}
\newcommand{\slsa}{s_{\mathrm{LSA}}}

\icmltitlerunning{Split Conformal Prediction with Label-Shift-Adjusted Bayesian Scores}

\begin{document}

\twocolumn[
\icmltitle{Split Conformal Prediction with Label-Shift-Adjusted Bayesian Scores}

\icmlsetsymbol{equal}{*}

\begin{icmlauthorlist}
\icmlauthor{Hyeonsu Lee}{equal,mogam}
\icmlauthor{Juyeon Kim}{equal,mogam}
\icmlauthor{Erkhembayar Jadamba}{equal,mogam}
\icmlauthor{Seungjin Choi}{croid}
\icmlauthor{Hyunjin Shin}{mogam}
\end{icmlauthorlist}

\icmlaffiliation{mogam}{MOGAM Institute for Biomedical Research, Korea}
\icmlaffiliation{croid}{CROID Research and aSSIST University, Korea}

\icmlcorrespondingauthor{Hyunjin Shin}{hyunjin.shin@mogam.re.kr}

\vskip 0.3in
]

\begin{abstract}
Conformal prediction provides distribution-free uncertainty quantification under exchangeability. However, this assumption is violated by label shift, where the marginal distribution of labels changes while the conditional distribution of inputs given labels remains stable. Under such shifts, standard conformal procedures no longer maintain their intended coverage behavior.
Existing approaches address this via importance weighting. They pair the
reweighting with residual-based nonconformity scores that ignore predictive
uncertainty. The resulting intervals have uniform width.
Bayesian conformal methods produce adaptive intervals by leveraging
predictive distributions. They evaluate conformity under the source
predictive, which is misaligned with the target domain under label shift. We propose the \emph{Label-Shift-Adjusted Bayesian Score} (LSA score), a
nonconformity score derived from a posterior predictive tilting identity.
This identity shows that the target predictive is an importance-weighted
transformation of the source predictive. We use it to derive a direct
correction to the Bayesian score.
We evaluate the method on molecular property prediction under controlled label shift. The LSA score consistently yields shorter intervals than residual-based and source-based Bayesian scores. Coverage in the target domain remains comparable. Under stronger shift, all methods incur some coverage loss due to pseudo-label-based density-ratio estimation. The LSA score is defined for any source predictive with a tractable log-density. We instantiate it with Bayesian Ridge Regression, where the correction admits a closed form.
\end{abstract}
\printAffiliationsAndNotice{\icmlEqualContribution}

\section{Introduction}\label{sec:intro}

Conformal prediction (CP) constructs prediction intervals with finite-sample
coverage guarantees under exchangeability
\citep{vovk2005algorithmic,shafer2008tutorial,angelopoulos2023conformal}.
This property makes it appealing for applications such as molecular property
prediction~\citep{laghuvarapu2023codrug}.
In practice, however, the exchangeability assumption is often violated by
\emph{label shift} \citep{saerens2002adjusting}. Under label shift, the marginal label
distribution changes while the conditional distribution of inputs given
labels remains the same.
This arises naturally in scientific settings when attention shifts toward
rare or extreme property values. It causes standard conformal methods to
lose their coverage guarantees.

A good prediction interval under label shift must satisfy two properties
simultaneously. First, the nonconformity score must be \emph{aligned} with
the target distribution so that the weighted quantile yields valid coverage.
Second, the score must be \emph{adaptive} to predictive uncertainty so that
the interval width reflects local difficulty. Prior work addresses these two
properties separately.

Importance-weighting methods target the first property. Tibshirani
et al.~\citep{tibshirani2019conformal} introduced weighted quantiles to
restore coverage under covariate shift. Barber
et al.~\citep{barber2023conformal} generalized this idea to a broader
framework that handles arbitrary violations of exchangeability. Podkopaev
and Ramdas~\citep{podkopaev2021distribution} and
Si et al.~\citep{si2024pac} extended the weighted approach to label shift
specifically. Lee et al.~\citep{lee2025conformal} applied it to molecular
property prediction. Gibbs and
Cand\`{e}s~\citep{gibbs2021adaptive} proposed online threshold adjustment
to track distribution drift over time. These methods reweight calibration
samples to match the target marginal. They restore coverage. However, the
nonconformity scores paired with the reweighting are residual-based. They
depend only on point predictions. The resulting intervals have uniform width
regardless of local difficulty. The second property is left unaddressed.

Adaptive-score methods target the second property. Romano
et al.~\citep{romano2019conformalized} proposed conformalized quantile
regression (CQR), which produces intervals whose width varies with input
difficulty under exchangeability. Fong and
Holmes~\citep{fong2021conformal} used the Bayesian posterior predictive as
a nonconformity score. The resulting intervals reflect model uncertainty.
Bhagwat et al.~\citep{bhagwat2025cbma} extended this idea through Bayesian
model averaging. Recent theoretical work by Datta
et al.~\citep{datta2025conformal} and Deliu and
Liseo~\citep{deliu2025interplay} further clarifies the relationship between
conformal prediction and Bayesian inference. These methods produce intervals
of varying width. However, they evaluate conformity under the source
predictive distribution. Under label shift, the source predictive is
misaligned with the target domain. This misalignment distorts the weighted
score distribution. The calibration quantile inflates and the intervals
become unnecessarily wide, as we verify empirically in
\cref{sec:quantile}. The first property is violated. No existing
nonconformity score satisfies both properties at once.

We address this gap by introducing the \textbf{Label-Shift-Adjusted Bayesian
(LSA) score}.
Our starting point is a posterior predictive tilting identity
(\cref{prop:tilting}). Under label shift, the target predictive
is an importance-weighted transformation of the source predictive.
The negative log-density of this identity yields a score decomposition that
directly motivates a correction to the source Bayesian score.
We instantiate the score with Bayesian Ridge Regression (BRR). The correction admits a
closed-form prediction interval with no additional computational cost.

We evaluate the method on molecular property prediction under controlled
label shift.
The LSA score yields shorter prediction intervals than residual and source
score baselines. Target-domain coverage remains comparable.
The method relies on pseudo-label-based density-ratio estimation, so we do
not claim exact finite-sample validity under shift.
We instead analyze the practical weighted quantile as an approximation to
an oracle procedure in \cref{sec:quantile}.
\section{Methods}\label{sec:methods}

\subsection{Problem Setup}\label{sec:setup}

Let $\bh\in\R^d$ denote a feature representation of an input, and let
$y\in\R$ be a scalar target variable. In our experiments, $\bh$ is a
molecular representation extracted by a pretrained encoder whose weights
are frozen, and $y$ is a molecular property such as aqueous solubility
($\log S$ in $\log\,\mathrm{mol/L}$).
Data are partitioned into four disjoint sets. The \emph{training set}
$\Dtrain=\{(\bh_i,y_i)\}_{i=1}^{n_{\mathrm{tr}}}$, \emph{weight
estimation set} $\Dweight=\{(\bh_i,y_i)\}_{i=1}^{n_{\mathrm{w}}}$, and
\emph{calibration set} $\Dcal=\{(\bh_i,y_i)\}_{i=1}^{n}$ are drawn from
the source distribution $\ps$. The \emph{target set}
$\Dtest=\{\bh_j\}_{j=1}^{m}$, observed without labels at test time, is
drawn from a shifted distribution $\pt$.

We model label shift via a one-parameter exponential tilt:
\begin{align}
  \pt(y) &\;\propto\; \ps(y)\,\exp(\beta y), \notag\\
  r(y) &:= \frac{\pt(y)}{\ps(y)}
  = \frac{\exp(\beta y)}{Z_r},
  \notag\\
  Z_r &:= \int \ps(y)\,\exp(\beta y)\,\mathrm{d}y,
  \label{eq:exp_tilt}
\end{align}
where $\beta\in\R$ controls the direction and magnitude of the shift.
$Z_r$ is the normalizing constant.
Label shift in general only requires $\pt(y)\neq\ps(y)$. We adopt the
exponential tilt as a tractable parametric model because it yields a
log-linear density ratio, which admits a closed-form correction under
Gaussian predictives (\cref{sec:lsa}).
Since the feature extractor is
deterministic, the label shift assumption carries over to the representation
level. That is, $\ps(\bh\mid y)=\pt(\bh\mid y)$.
Our goal is to construct prediction intervals
$C(\bh)=[L(\bh),\,U(\bh)]$ whose target-domain marginal coverage
$\mathbb{P}_{(\bh,y)\sim \pt}(y\in C(\bh))$ is close to $1-\acp$.
Because our implementation relies on estimated density ratios obtained from
pseudo-labels, we do not claim exact finite-sample validity under shift.

\subsection{Posterior Predictive Tilting}\label{sec:tilting}

Under label shift, the source predictive $p_s(y\mid\bh,\Dtrain)$ is no
longer aligned with the target domain. A natural question is whether the
target posterior predictive can be recovered from the source one without
retraining. We show that it can, via a simple importance-weighting identity.
The following assumptions formalize the required conditions.

\begin{assumption}[Conditional label shift after representation]
\label{asm:label_shift}
For all $\bh\in\mathcal{H}$ and $y\in\mathcal{Y}$,
\[
  \ps(\bh\mid y)=\pt(\bh\mid y).
\]
We further assume the corresponding predictive-level invariance after
source-model fitting:
\[
  \ps(\bh\mid y,\Dtrain)=\pt(\bh\mid y,\Dtrain).
\]
\end{assumption}

\begin{assumption}[Absolute continuity]
\label{asm:abs_cont}
$\ps(y)>0$ wherever $\pt(y)>0$, so that
$r(y)=\pt(y)/\ps(y)$ is well-defined.
\end{assumption}

\begin{assumption}[Source representativeness]
\label{asm:source_rep}
The source predictive model is well-specified and the posterior concentrates, so that
\[
  \ps(y\mid \Dtrain)=\ps(y).
\]
This is a mild condition satisfied whenever the training sample is reasonably large.
\end{assumption}

\begin{proposition}[Posterior predictive tilting]
\label{prop:tilting}
Under \cref{asm:label_shift}, \cref{asm:abs_cont}, and
\cref{asm:source_rep}, and conditioning on the fitted source data $\Dtrain$,
the target posterior predictive admits the representation
\begin{equation}\label{eq:tilting}
  p_t(y \mid \bh, \Dtrain)
  = \frac{p_s(y \mid \bh, \Dtrain)\, r(y)}{Z(\bh,\Dtrain)},
\end{equation}
where
\begin{equation}
\begin{aligned}
Z(\bh,\Dtrain)
&:=
\int p_s(y' \mid \bh,\Dtrain)\, r(y')\, dy'\\
&=
\mathbb{E}_{Y \sim p_s(\cdot \mid \bh,\Dtrain)}\![r(Y)].
\end{aligned}
\label{eq:Z_def}
\end{equation}
\end{proposition}

The derivation is given in \cref{app:proof}. It applies Bayes' rule under the
$\Dtrain$-conditioned predictive distributions. It uses \cref{asm:label_shift}
to replace the target conditional mechanism by its source counterpart. It then
identifies the marginal ratio through \cref{asm:source_rep}. Since $\Dtrain$ is drawn entirely from the source domain, it carries no
information about the target label marginal, so
$\pt(y\mid\Dtrain) = \pt(y)$. Combined with \cref{asm:source_rep},
the conditional marginal ratio reduces to the population density ratio
$r(y) = \pt(y)/\ps(y)$.

This identity has a direct implication for nonconformity scoring.
Using $s(\bh,y) = -\log p(y\mid\bh,\Dtrain)$ under the source predictive
yields a score misaligned with the target domain. \Cref{prop:tilting} shows
that the correctly aligned score uses $p_t$ instead of $p_s$. Taking the
negative logarithm of \cref{eq:tilting} yields
\begin{multline}
  -\log \pt(y\mid\bh,\Dtrain)
  =
  \underbrace{-\log \ps(y\mid\bh,\Dtrain)}_{\text{source score}}
  \\
  +
  \underbrace{(-\log r(y))}_{\text{shift correction}}
  +
  \underbrace{\log Z(\bh,\Dtrain)}_{\text{normalization}}.
  \label{eq:score_decomp}
\end{multline}
Define the oracle nonconformity score as
$s_t(\bh,y) := -\log \pt(y\mid\bh,\Dtrain)$.
The sublevel set $\{y : s_t(\bh,y) \le q\}$ coincides exactly with the
highest-density region of the target predictive at level $e^{-q}$.
This decomposition motivates our practical score construction.

\subsection{Label-Shift-Adjusted Bayesian Score}\label{sec:lsa}

Guided by \cref{eq:score_decomp}, we first estimate the density ratio
$r(y)$. Label shift estimation has been studied extensively
\citep{lipton2018detecting,saerens2002adjusting}. We use the estimated
ratio to define the target-aligned predictive and the corresponding
nonconformity score.
Under the exponential tilt model (\cref{eq:exp_tilt}), the log-density ratio
is exactly linear in $y$. We estimate it by fitting a logistic regression
classifier~\citep{sugiyama2012density} that distinguishes source labels
$\{y_i\}_{i\in\Dweight}$ from
target pseudo-labels $\{\tilde y_j\}_{j\in\Dtest}$, where
\begin{equation}
  \tilde y_j := \mu(\bh_j)
  \label{eq:pseudo_label}
\end{equation}
is the predictive mean of the source model at each unlabeled target input
(defined in \cref{eq:pred_dist} below).
This yields
\begin{equation}
  \log \hat r(y)
  =
  \hat\beta_0 + \hat\beta_1 y,
  \label{eq:est_weights}
\end{equation}
where $\hat\beta_0$ absorbs the normalization constant and the class-prior
offset (see \cref{app:weight_relation}).
With $\hat r(y)$ in hand, we replace $r(y)$ in \cref{eq:tilting} with
$\hat r(y)$ and define the \emph{estimated target-aligned predictive}
\begin{align}
  \hat p_t(y\mid\bh,\Dtrain)
  &:=
  \frac{\ps(y\mid\bh,\Dtrain)\,\hat r(y)}
       {\hat Z(\bh)},
  \notag\\
  \hat Z(\bh)
  &:=
  \int \ps(y'\mid\bh,\Dtrain)\,\hat r(y')\,\mathrm{d}y'.
  \label{eq:pt_tilt_hat}
\end{align}

\paragraph{LSA score.}
Dropping the constant $\frac{1}{2}\log(2\pi)$ throughout, the source score
is the negative log-density of $\ps(y\mid\bh,\Dtrain)$,
\begin{equation}
  s_{\mathrm{s}}(\bh,y)
  =
  \frac{(y-\mu(\bh))^2}{2\sigma^2(\bh)}
  +
  \frac{1}{2}\log\sigma^2(\bh),
  \label{eq:s_score}
\end{equation}
and we define the \textbf{Label-Shift-Adjusted Bayesian Score} (LSA score) as
\begin{equation}
  \slsa(\bh,y)
  =
  s_{\mathrm{s}}(\bh,y)
  - \log \hat r(y)
  + \log \hat Z(\bh).
  \label{eq:s_lsa}
\end{equation}
This is the negative log-density of $\hat p_t(y\mid\bh,\Dtrain)$ up to the
dropped constant. It is therefore a plug-in approximation to the oracle
target score in \cref{eq:score_decomp}, with $\hat r(y)$ in place of $r(y)$.
The factor $\hat r(y)$ inside the score reshapes what is measured. It replaces
source-based conformity with conformity under the estimated target-aligned
predictive. Specifically, $-\log\hat r(y)$ lowers the nonconformity score for
target-favored labels and raises it for target-disfavored ones.
$\log\hat Z(\bh)$ restores normalization through an input-dependent offset.
Note that this score-level correction is distinct from the weight factor
$\hat r(y_i)$ used later to reweight calibration scores in the weighted
quantile. The score correction changes what is measured. It evaluates
conformity under the target-aligned predictive rather than the source one.
The weight factor changes how calibration samples are aggregated. It
reweights the empirical distribution to match the target marginal.
The two address different axes of the label shift problem and should not be
conflated.

The LSA score is defined for any source predictive $p_s(y\mid\bh,\Dtrain)$.
We instantiate it with Bayesian Ridge Regression (BRR). The Gaussian
predictive and log-linear density ratio together yield a closed-form
prediction interval with no additional computational cost over the source
score.
BRR places a Gaussian linear model on the input representations:
\begin{equation}
  y = \bh^\top\bw + \varepsilon,
  \quad
  \varepsilon\sim\Ncal(0,\,\lambda^{-1}),
  \quad
  \bw\sim\Ncal(\bm{0},\,\alpha^{-1}\bI_d),
  \label{eq:brr_model}
\end{equation}
where hyperparameters are estimated by marginal likelihood maximization
\citep{tipping2001sparse,bishop2006pattern} (see~\cref{app:brr_posterior}).
Conjugacy yields a Gaussian posterior predictive
\begin{align}
  \ps(y^*\mid\bh^*,\Dtrain)
    &= \Ncal\!\bigl(y^*\mid
      \mu(\bh^*),\;\sigma^2(\bh^*)\bigr),
  \notag\\
  \mu(\bh^*)&={\bh^*}^\top\bmu_{\bw},
  \notag\\
  \sigma^2(\bh^*)&=
      {\bh^*}^\top\bSigma_{\bw}\bh^*+\lambda^{-1},
  \label{eq:pred_dist}
\end{align}
Because $\ps(y\mid\bh,\Dtrain)$ is Gaussian and $\hat r(y)$ is log-linear
in $y$, both $\hat Z(\bh)$ and the tilted predictive admit analytic forms
(derivations in \cref{app:Z_derivation,app:gaussian_tilt}):
\begin{align}
  \hat Z(\bh)
  &=
  \exp\!\Bigl(
    \hat\beta_0
    + \hat\beta_1\,\mu(\bh)
    + \tfrac{1}{2}\hat\beta_1^2\,\sigma^2(\bh)
  \Bigr),
  \label{eq:Z_analytic}\\
  \hat p_t(y \mid \bh,\Dtrain)
  &=
  \mathcal{N}\!\bigl(y \mid \mu^*(\bh), \sigma^2(\bh)\bigr),
  \notag\\
  \mu^*(\bh)&:=\mu(\bh)+\hat\beta_1\sigma^2(\bh),
  \label{eq:tilted_gaussian}
\end{align}
so exponential tilting shifts only the predictive mean. The variance remains
unchanged. The LSA score therefore reduces to
\begin{equation}
  s_{\mathrm{LSA}}(\bh,y)
  =
  \frac{(y-\mu^*(\bh))^2}{2\sigma^2(\bh)}
  +\frac12\log \sigma^2(\bh),
  \label{eq:lsa_gaussian_score}
\end{equation}
which coincides with the exact target score $-\log p_t(y\mid\bh,\Dtrain)$
of \cref{prop:tilting} when $\hat r(y)=r(y)$.

\subsection{Weighted Conformal Prediction with the LSA Score}

Given calibration scores $\{s_{\mathrm{LSA}}(\bh_i, y_i)\}_{i=1}^n$ and
estimated density ratios $\{\hat r(y_i)\}_{i=1}^n$, we compute the
weighted empirical quantile following
Tibshirani et al.~\citep{tibshirani2019conformal}:
\begin{multline}
\hat q
=
\inf\Bigl\{
q :
\frac{\sum_{i=1}^n \hat r(y_i)\,\mathbf{1}[s_{\mathrm{LSA}}(\bh_i,y_i)\le q]}
     {\sum_{i=1}^n \hat r(y_i) + \hat r(\tilde y_{n+1})}\\
\ge
1-\alpha_{\mathrm{cp}}
\Bigr\},
\label{eq:weighted_quantile}
\end{multline}
where $\tilde y_{n+1}=\mu(\bh_{n+1})$ is the pseudo-label of the test
input. Since the true test label is unobserved, we use the predictive mean
as a surrogate. This is consistent with the pseudo-label construction
in~\cref{eq:pseudo_label}.

The LSA score takes the Gaussian log-density form in
\cref{eq:lsa_gaussian_score}. The prediction interval for a test input
$\bh^\ast$ is
\begin{equation}
C(\bh^\ast)
=
\Bigl[\mu^\ast(\bh^\ast)\pm
\sigma(\bh^\ast)\sqrt{2\max\bigl(
  \hat q-\tfrac{1}{2}\log\sigma^2(\bh^\ast),0\bigr)}\,\Bigr],
\label{eq:clipped_interval}
\end{equation}
centered at the label-shift-adjusted mean $\mu^\ast(\bh^\ast)$ with
half-width proportional to $\sigma(\bh^\ast)$. The intervals are therefore
sample-adaptive (derivation in \cref{app:score_derivation}).
The primary efficiency gain stems from reshaping the score distribution.
The LSA correction shifts the predictive center to
$\mu^\ast(\bh)=\mu(\bh)+\hat\beta_1\sigma^2(\bh)$. Target-favored labels
therefore receive lower nonconformity scores. This shifts the weighted CDF
leftward and lowers $\hat q$. We evaluate this effect empirically in
\cref{sec:experiments}.

\section{Experiments}\label{sec:experiments}
\subsection{Experimental Setup}\label{sec:exp_setup}

We evaluate on AqSolDB from the
Therapeutics Data Commons~\citep{sorkun2019aqsoldb,Huang2021tdc},
which contains $9{,}982$ compounds with experimentally measured aqueous
solubility ($\log S$ in $\log\,\mathrm{mol/L}$). Molecular SMILES strings
are encoded with a frozen pretrained BART-based chemical language
model~\citep{ross2022large} ($d=768$, $171.16$M parameters). Only the
BRR head is fitted on the extracted representations
(see~\cref{app:encoder_details} for pretraining details).
Results on a second benchmark, Lipophilicity~\citep{wenlock2015experimental,wu2018moleculenet},
are reported in~\cref{app:lipophilicity_robustness}.

In each trial, the dataset is randomly partitioned into four disjoint
subsets. $\Dtrain$ ($\sim$40\%) is used for model training. Three
equal-sized subsets (each 20\%) serve as $\Dweight$ for density-ratio
estimation, $\Dcal$ for conformal calibration, and $\Dtest$ for evaluation.
For shifted evaluation, $\Dtest$ is resampled with replacement using
probabilities proportional to $\exp(\beta y)$, following the
exponential-tilt scheme in~\cref{eq:exp_tilt} with
$\beta\in\{0.0,\,-0.1,\,-0.2,\,-0.3,\,-0.4,\,-0.5\}$.
Here $\beta=0$ corresponds to no shift and $\beta=-0.5$ to a moderate
shift toward low-solubility compounds.
Each configuration is repeated over 1,000 random seeds. We report
marginal coverage $\hat P(y \in C(\bh))$ at target level $0.9$ and
average interval length $\E[U(\bh)-L(\bh)]$.

All methods share the same BRR predictive model, logistic regression
weight estimator, and weighted quantile computation (\cref{eq:weighted_quantile}).
They differ only in the nonconformity score.
\textbf{Residual} uses $s_{\mathrm{res}}(\bh,y)=|y-\mu(\bh)|$ and produces
uniform intervals. \textbf{Source score} uses $s_{\mathrm{s}}(\bh,y)$
in~\cref{eq:s_score} and produces adaptive intervals without label-shift
correction. \textbf{LSA} uses $\slsa(\bh,y)$ in~\cref{eq:s_lsa} and
produces adaptive intervals with label-shift correction.
This controlled comparison isolates the effect of the conformity score.

\subsection{Main Results}\label{sec:main_results}

\begin{figure*}[!t]
\centering
\includegraphics[width=\textwidth]{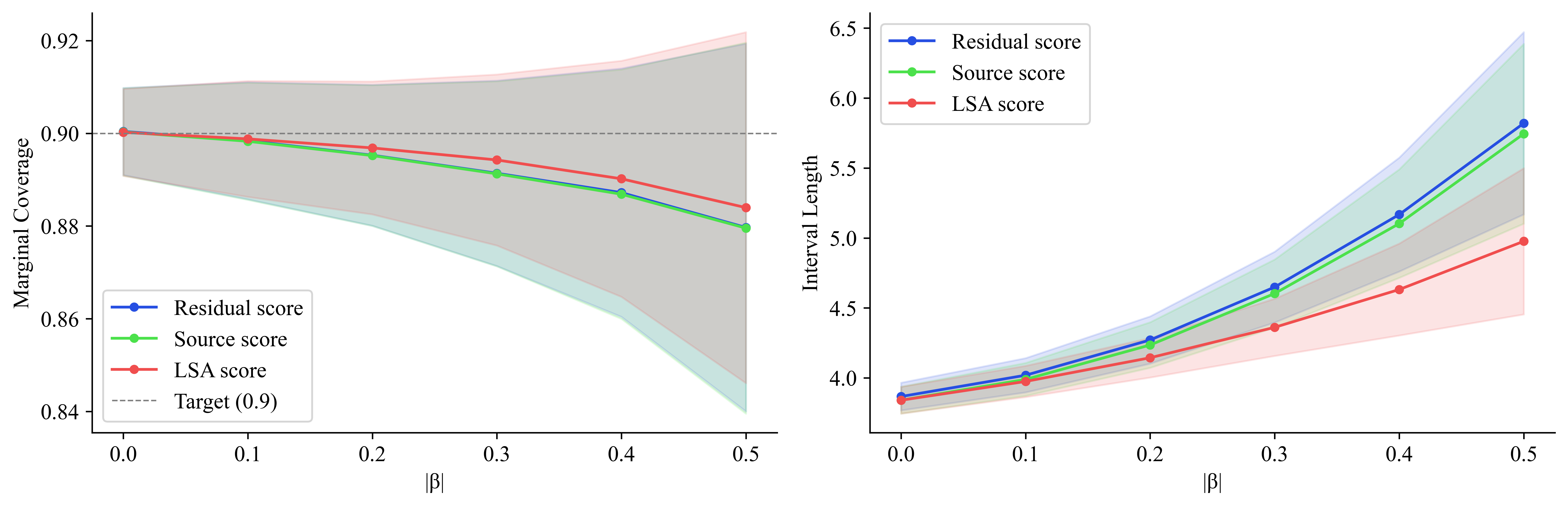}
\caption{Coverage (left) and interval length (right) under exponential-tilt
label shift on AqSolDB.  Three nonconformity scores are compared. Residual,
Source score, and LSA are shown.  Target coverage is $0.9$.
$\beta\in\{0,\,-0.1,\ldots,\,-0.5\}$.  Error bands denote $\pm$1 std over
$1{,}000$ seeds.}
\label{fig:main}
\end{figure*}

\Cref{tab:main,fig:main} summarize the main results. At $\beta=0$, all three scores achieve the nominal $90\%$ coverage with nearly identical interval lengths. This confirms correct calibration under exchangeability. As $|\beta|$ increases, the LSA score yields shorter intervals than both baselines. The gap widens with $|\beta|$. At $\beta=-0.5$, the average interval length is $4.98$ for LSA, compared to $5.74$ for Source score and $5.82$ for Residual, corresponding to reductions of $13\%$ and $14\%$, respectively. Paired $t$-tests on interval length yield $p<0.001$ for all $|\beta|\geq 0.1$. This improvement reflects a smaller weighted score quantile under LSA. We analyze this through the weighted CDF in \cref{sec:quantile}.

We observe the same qualitative behavior under an alternative Gaussian-based density-ratio estimator (\cref{app:gaussian_mean}). This is consistent with the shared log-linear form in $y$. The log-linear form preserves the Gaussian structure of the LSA correction.
All methods lose coverage as $|\beta|$ increases. This is expected because the weighted quantile is computed from pseudo-label-based weight estimates rather than oracle importance weights. As the shift becomes stronger, the target pseudo-labels $\tilde y_j$ used for density-ratio estimation become less accurate. This degrades the estimated importance weights and the resulting weighted conformal quantile. We therefore interpret the remaining coverage gap primarily as an estimation effect rather than as evidence against the target-aligned score itself. Within this estimated-weight regime, LSA remains competitive in coverage. It achieves substantially shorter intervals.

\begin{table}[!tbp]
\caption{Coverage and average interval length under exponential-tilt label
shift. Target coverage is $0.9$. Results are averaged over $1{,}000$ seeds.
\textbf{Bold} indicates the best value for each $\beta$.}
\label{tab:main}
\centering
\footnotesize
\setlength{\tabcolsep}{4pt}
\begin{tabular}{@{}l ccc@{}}
\toprule
\multicolumn{4}{c}{Coverage $\uparrow$} \\
\cmidrule(lr){1-4}
$\beta$ & Res. & Source score & LSA \\
\midrule
$0.0$  & $\mathbf{.9004_{\pm.0094}}$ & $.9003_{\pm.0094}$ & $.9002_{\pm.0094}$ \\
$-0.1$ & $.8984_{\pm.0126}$ & $.8983_{\pm.0126}$ & $\mathbf{.8988_{\pm.0125}}$ \\
$-0.2$ & $.8953_{\pm.0152}$ & $.8952_{\pm.0152}$ & $\mathbf{.8969_{\pm.0143}}$ \\
$-0.3$ & $.8914_{\pm.0200}$ & $.8913_{\pm.0200}$ & $\mathbf{.8943_{\pm.0185}}$ \\
$-0.4$ & $.8872_{\pm.0268}$ & $.8869_{\pm.0269}$ & $\mathbf{.8902_{\pm.0255}}$ \\
$-0.5$ & $.8797_{\pm.0397}$ & $.8795_{\pm.0400}$ & $\mathbf{.8842_{\pm.0378}}$ \\
\midrule
\multicolumn{4}{c}{Interval Length $\downarrow$} \\
\cmidrule(lr){1-4}
$\beta$ & Res. & Source score & LSA \\
\midrule
$0.0$  & $3.867_{\pm.099}$ & $3.841_{\pm.096}$ & $\mathbf{3.840_{\pm.096}}$ \\
$-0.1$ & $4.019_{\pm.122}$ & $3.989_{\pm.118}$ & $\mathbf{3.975_{\pm.112}}$ \\
$-0.2$ & $4.271_{\pm.169}$ & $4.235_{\pm.163}$ & $\mathbf{4.144_{\pm.141}}$ \\
$-0.3$ & $4.649_{\pm.253}$ & $4.603_{\pm.244}$ & $\mathbf{4.362_{\pm.204}}$ \\
$-0.4$ & $5.168_{\pm.407}$ & $5.103_{\pm.390}$ & $\mathbf{4.632_{\pm.329}}$ \\
$-0.5$ & $5.820_{\pm.651}$ & $5.744_{\pm.640}$ & $\mathbf{4.977_{\pm.524}}$ \\
\bottomrule
\end{tabular}
\end{table}

\subsection{Score Distribution Shift Analysis}\label{sec:quantile}

\begin{figure*}[!t]
  \centering
  \includegraphics[width=\textwidth]{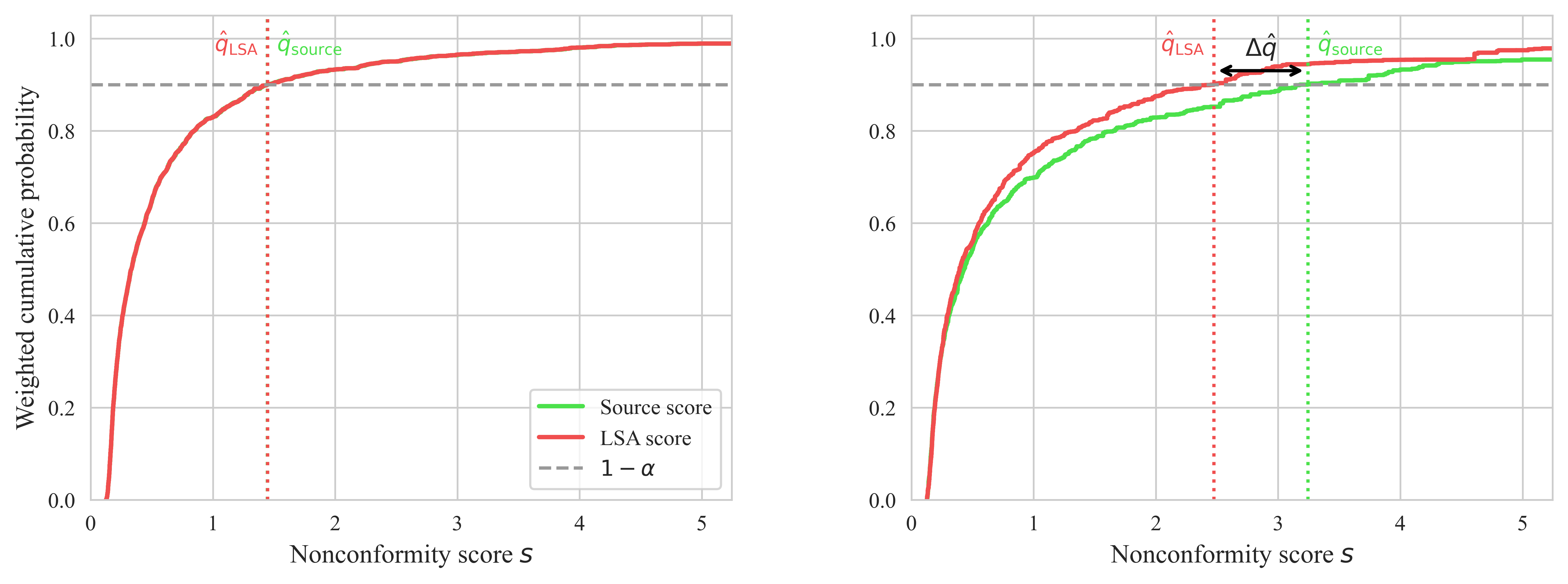}
  \caption{Weighted CDFs of calibration scores at $\beta=0.0$ (left) and
$\beta=-0.5$ (right) on AqSolDB.  Each curve plots the cumulative weight
fraction $\sum \hat r(y_i)\mathbf{1}[s_i\le q]\,/\,\sum \hat r(y_j)$ as a
function of the score threshold $q$.  The weighted quantile $\hat q$ is the
score at which the CDF reaches $1-\acp\approx 0.9$.
At $\beta=0.0$, the Source score and LSA CDFs nearly overlap. At
$\beta=-0.5$, the LSA CDF shifts leftward and yields a smaller weighted
quantile.}
  \label{fig:wcdf}
\end{figure*}

\Cref{fig:wcdf} compares the weighted CDFs of the Source score and LSA calibration scores.
The weighted quantile $\hat q$ is the score at which the CDF reaches
$1-\acp \approx 0.9$. A smaller quantile yields a shorter interval
through~\cref{eq:clipped_interval}. At $\beta=0$, the two CDFs nearly overlap.
Their quantiles coincide. At $\beta=-0.5$, the LSA CDF shifts leftward and
the quantile decreases. This follows from the mean shift in~\cref{eq:tilted_gaussian}.
The LSA score evaluates conformity with respect to the estimated target-aligned
mean $\mu^*(\bh)=\mu(\bh)+\hat\beta_1\sigma^2(\bh)$ rather than the source mean
$\mu(\bh)$. Under negative label shift ($\hat\beta_1 < 0$), calibration samples
with target-favored labels receive lower LSA scores than Source scores. This
shifts weighted mass leftward and lowers $\hat q$. The Source score evaluates
conformity under the source predictive and does not perform this correction. It
therefore assigns higher scores to target-favored samples. This produces a larger
weighted quantile and wider intervals.

This score distribution shift directly explains the interval-length reduction
observed in \cref{tab:main,fig:main}. Coverage remains broadly similar across
methods within the estimated-weight regime. All methods degrade under stronger
shift. A likely reason is that stronger label shift makes the target
pseudo-labels $\tilde y_j$ used for density-ratio estimation less reliable.
This directly affects the estimated density ratios and the resulting weighted
conformal quantile. In this regime, the re-centering in LSA improves efficiency
without introducing an additional coverage penalty.

To assess whether this
effect is specific to AqSolDB, we evaluate the same protocol on a second
molecular regression benchmark, Lipophilicity
(see~\cref{app:lipophilicity_robustness}). At $\beta=0$, the three methods are
nearly indistinguishable. As the magnitude of negative label shift increases,
the LSA score again yields shorter intervals than both baselines. Empirical
coverage remains comparable and is slightly higher for LSA under larger shifts.

\section{Conclusion}\label{sec:conclusion}
We studied conformal prediction under label shift and identified a key
limitation of existing approaches. Importance weighting can partially
correct distribution shift. Commonly used nonconformity scores, however,
ignore predictive uncertainty. This leads to inefficient prediction intervals.
We proposed the Label-Shift-Adjusted Bayesian Score, a nonconformity score
derived from a posterior predictive tilting identity. Label shift induces a
simple transformation of the posterior predictive. This motivates a direct
correction to the standard Bayesian score. Under Bayesian Ridge
Regression, the method yields closed-form, sample-adaptive prediction
intervals with negligible computational overhead.

Empirically, the LSA score produces shorter prediction intervals than both residual-based and source-based Bayesian scores. Coverage remains comparable across a range of label shift magnitudes. As the shift becomes more pronounced, all methods exhibit some loss in coverage. This behavior is consistent with the use of pseudo-labels for density-ratio estimation. Pseudo-labels introduce additional approximation error. Within this regime, the improvement from LSA can be traced to a better alignment between the score distribution and the target predictive.

This study uses a Gaussian predictive model. The adjustment appears as a shift in the predictive mean. The same idea can be applied more broadly. Extending the approach to richer models such as deep ensembles or sampling-based methods (MC-Dropout) is a natural next step.

\section*{Acknowledgements}
This research was supported by a grant of the Korea Machine Learning Ledger Orchestration for Drug Discovery Project (K-MELLODDY), funded by the Ministry of Health \& Welfare and Ministry of Science and ICT, Republic of Korea (grant number: RS-2024-00459964).

\bibliographystyle{plainnat}
\bibliography{references}

\clearpage
\onecolumn
\appendix

\section{BRR Posterior and Predictive Distribution}
\label{app:brr_posterior}

Conjugacy yields the closed-form posterior
$p(\bw\mid\Dtrain)=\Ncal(\bw\mid\bmu_{\bw},\bSigma_{\bw})$ with
\begin{align}
  \bSigma_{\bw}
    &= \bigl(\alpha\,\bI_d + \lambda\,\bH^\top\bH\bigr)^{-1},
  \label{eq:post_cov}\\
  \bmu_{\bw}
    &= \lambda\,\bSigma_{\bw}\,\bH^\top\by,
  \label{eq:post_mean}
\end{align}
where $\bH\in\R^{n_{\mathrm{tr}}\times d}$ stacks the training
representations row-wise and $\by\in\R^{n_{\mathrm{tr}}}$ collects labels.
The predictive mean and variance in~\cref{eq:pred_dist}
follow by marginalizing over $\bw$.

\section{Full Proof of \cref{prop:tilting}}
\label{app:proof}

We provide the complete derivation, conditioning throughout on the fitted
source dataset $\Dtrain$.

By Bayes' rule in the target domain,
\begin{equation}
  \pt(y\mid\bh,\Dtrain)
  =
  \frac{\pt(\bh\mid y,\Dtrain)\;\pt(y|\Dtrain)}
       {\pt(\bh\mid\Dtrain)}.
  \label{eq:app_bayes}
\end{equation}
By ~\cref{asm:label_shift},

\[
  \pt(\bh\mid y,\Dtrain)=\ps(\bh\mid y,\Dtrain).
\]
Substituting this into~\cref{eq:app_bayes} gives
\begin{equation}
  \pt(y\mid\bh,\Dtrain)
  =
  \frac{\ps(\bh\mid y,\Dtrain)\;\pt(y|\Dtrain)}
       {\pt(\bh\mid\Dtrain)}.
  \label{eq:app_sub1}
\end{equation}

Applying Bayes' rule in the source domain,
\begin{equation}
  \ps(\bh\mid y,\Dtrain)
  =
  \frac{\ps(y\mid\bh,\Dtrain)\;\ps(\bh\mid\Dtrain)}
       {\ps(y|\Dtrain)}.
  \label{eq:app_source_bayes}
\end{equation}
Inserting~\cref{eq:app_source_bayes} into~\cref{eq:app_sub1}, we obtain
\begin{align}
  \pt(y\mid\bh,\Dtrain)
  &=
  \ps(y\mid\bh,\Dtrain)\cdot
  \frac{\pt(y|\Dtrain)}{\ps(y|\Dtrain)}\notag\\
  &\quad\cdot
  \frac{\ps(\bh\mid\Dtrain)}{\pt(\bh\mid\Dtrain)}.
  \label{eq:app_combined}
\end{align}

\paragraph{Identifying the label-marginal ratio.}
At this point, the derivation introduces the conditional label-marginal ratio
$\pt(y\mid\Dtrain)/\ps(y\mid\Dtrain)$.
Since $\Dtrain$ is drawn entirely from the source domain $\ps$, it carries
no information about the target label distribution. We therefore have
$\pt(y\mid\Dtrain) = \pt(y)$. By \cref{asm:source_rep},
$\ps(y\mid\Dtrain) = \ps(y)$. It follows that
\begin{equation}
  \frac{\pt(y\mid\Dtrain)}{\ps(y\mid\Dtrain)}
  =
  \frac{\pt(y)}{\ps(y)}
  =
  r(y).
  \label{eq:app_r_bridge}
\end{equation}

Using $r(y)$, we rewrite~\cref{eq:app_combined} as
\begin{equation}
  \pt(y\mid\bh,\Dtrain)
  =
  \ps(y\mid\bh,\Dtrain)\,r(y)\,
  \frac{\ps(\bh\mid\Dtrain)}{\pt(\bh\mid\Dtrain)}.
  \label{eq:app_combined_r}
\end{equation}

\paragraph{Identifying $1/Z(\bh,\Dtrain)$.}
Using the law of total probability, Bayes' rule, and \cref{asm:label_shift},
\begin{align}
  \pt(\bh\mid\Dtrain)
  &= \int \pt(\bh\mid y',\Dtrain)\;\pt(y')\;\mathrm{d}y'
  \notag\\
  &= \int \ps(\bh\mid y',\Dtrain)\;\pt(y')\;\mathrm{d}y'
  \notag\\
  &= \int
     \frac{\ps(y'\mid\bh,\Dtrain)\;\ps(\bh\mid\Dtrain)}
          {\ps(y')}\;\pt(y')\;\mathrm{d}y'
  \notag\\
  &= \ps(\bh\mid\Dtrain)
     \int \ps(y'\mid\bh,\Dtrain)\,r(y')\,\mathrm{d}y'.
  \label{eq:app_Z_step}
\end{align}
Hence, with
\begin{align}
  Z(\bh,\Dtrain)
  &:=
  \int \ps(y'\mid\bh,\Dtrain)\,r(y')\,\mathrm{d}y' \notag\\
  &=
  \E_{Y\sim\ps(\,\cdot\,\mid\bh,\Dtrain)}\!\bigl[r(Y)\bigr],
  \label{eq:app_Z}
\end{align}
we have
\begin{align}
  \pt(\bh\mid\Dtrain)
  &=
  \ps(\bh\mid\Dtrain)\,Z(\bh,\Dtrain), \notag\\
  \frac{\ps(\bh\mid\Dtrain)}{\pt(\bh\mid\Dtrain)}
  &=
  \frac{1}{Z(\bh,\Dtrain)}.
  \label{eq:app_Z_ratio}
\end{align}
Substituting~\cref{eq:app_Z_ratio} into~\cref{eq:app_combined_r} yields
\begin{equation}
  \pt(y\mid\bh,\Dtrain)
  =
  \frac{\ps(y\mid\bh,\Dtrain)\,r(y)}
       {Z(\bh,\Dtrain)},
  \label{eq:app_final_tilting}
\end{equation}
which proves ~\cref{prop:tilting}.

\section{Population Tilt and the Estimated Log-Density Ratio}
\label{app:weight_relation}

Under the exponential tilt model,
\begin{equation}
  \pt(y)
  \propto
  \ps(y)\exp(\beta y),
  \label{eq:app_tilt_model}
\end{equation}
the population density ratio is
\begin{align}
  r(y)
  &=
  \frac{\pt(y)}{\ps(y)}
  =
  \frac{\exp(\beta y)}{Z_r}, \notag\\
  Z_r &:= \int \ps(y)\,\exp(\beta y)\,\mathrm{d}y.
  \label{eq:app_ratio_def}
\end{align}
Therefore, the log-density ratio is exactly linear in $y$:
\begin{equation}
  \log r(y)
  =
  \beta_0 + \beta_1 y,
  \qquad
  \beta_1 = \beta,
  \qquad
  \beta_0 = -\log Z_r.
  \label{eq:app_pop_to_est}
\end{equation}

If $\ps(y)$ is also Gaussian,
$\ps(y)=\Ncal(\mu_s,\sigma_s^2)$, then the moment-generating function gives
\begin{equation}
  Z_r
  =
  \exp\Bigl(\beta\mu_s+\tfrac{1}{2}\beta^2\sigma_s^2\Bigr),
  \label{eq:app_Zw_gaussian}
\end{equation}
so that
\begin{equation}
  \beta_0
  =
  -\beta\mu_s-\tfrac{1}{2}\beta^2\sigma_s^2.
  \label{eq:app_beta0_gaussian}
\end{equation}
This Gaussian calculation is only an illustrative expansion of the intercept
$\beta_0$. The log-linearity in~\cref{eq:app_pop_to_est} follows directly
from the exponential tilt model and does not require Gaussianity of $\ps(y)$.

In practice, we estimate the log-linear form in~\cref{eq:app_pop_to_est} by
fitting a logistic regression classifier that distinguishes source labels from
target pseudo-labels. This yields
\begin{equation}
  \log \hat r(y)
  =
  \hat\beta_0 + \hat\beta_1 y.
  \label{eq:app_est_ratio}
\end{equation}
The target sample is unlabeled. The pseudo-labels are noisy surrogates
for the true target labels. $\hat r(y)$ should therefore be viewed as a
practical approximation to the population density ratio, not an oracle
estimator. We use this estimated ratio throughout the LSA score construction.

\section{Closed-Form Derivation of \texorpdfstring{$\hat Z(\bh)$}{Z(h)} and the Tilted Predictive}
\label{app:Z_derivation}

Under the BRR model,

\[
  \ps(y\mid\bh,\Dtrain)=\Ncal(y\mid\mu(\bh),\sigma^2(\bh)),
\]
and the estimated density ratio is

\[
  \hat r(y)=\exp(\hat\beta_0+\hat\beta_1 y).
\]
The estimated normalizing constant is therefore
\begin{align}
  \hat Z(\bh)
  &= \int \ps(y\mid\bh,\Dtrain)\,\hat r(y)\,\mathrm{d}y
  \notag\\
  &= \int \Ncal(y\mid\mu(\bh),\sigma^2(\bh))\,
     \exp(\hat\beta_0+\hat\beta_1 y)\,\mathrm{d}y
  \notag\\
  &= \exp(\hat\beta_0)\,
     \E_{Y\sim\Ncal(\mu(\bh),\sigma^2(\bh))}\!
     \bigl[\exp(\hat\beta_1 Y)\bigr]
  \notag\\
  &= \exp(\hat\beta_0)\,
     \exp\!\Bigl(
       \hat\beta_1\mu(\bh)
       + \tfrac{1}{2}\hat\beta_1^2\sigma^2(\bh)
     \Bigr)
  \notag\\
  &= \exp\!\Bigl(
     \hat\beta_0
     + \hat\beta_1\mu(\bh)
     + \tfrac{1}{2}\hat\beta_1^2\sigma^2(\bh)
     \Bigr),
  \label{eq:app_Z_closed}
\end{align}
where the third equality uses the moment-generating function of the Gaussian:

\[
  \E[\exp(tY)]
  =
  \exp\!\Bigl(t\mu+\tfrac{1}{2}t^2\sigma^2\Bigr),
  \qquad
  Y\sim\Ncal(\mu,\sigma^2).
\]

\section{Gaussian Form of the Tilted Predictive}
\label{app:gaussian_tilt}

Starting from \cref{eq:pt_tilt_hat}, under the BRR predictive
\[
  \ps(y\mid\bh,\Dtrain)=\Ncal(y\mid \mu(\bh),\sigma^2(\bh))
\]
and the log-linear ratio model
\[
  \hat r(y)=\exp(\hat\beta_0+\hat\beta_1 y),
\]
we have
\begin{align}
  \hat p_t(y\mid\bh,\Dtrain)
  &\propto
  \Ncal(y\mid \mu(\bh),\sigma^2(\bh))
  \exp(\hat\beta_0+\hat\beta_1 y)\\
  &= \frac{1}{\sqrt{2\pi\sigma^2(\bh)}}
     \exp\!\Bigl(-\frac{(y-\mu(\bh))^2}{2\sigma^2(\bh)}
     + \hat\beta_0\notag\\
  &\qquad\qquad\qquad\qquad\qquad + \hat\beta_1 y\Bigr). \notag
\end{align}
Completing the square in the exponent with respect to $y$,
\begin{align}
  -\frac{(y-\mu(\bh))^2}{2\sigma^2(\bh)} + \hat\beta_1 y
  &=
  -\frac{y^2-2\mu(\bh)y+\mu(\bh)^2}{2\sigma^2(\bh)}
  + \hat\beta_1 y
  \notag\\
  &=
  -\frac{y^2-2\bigl(\mu(\bh)+\hat\beta_1\sigma^2(\bh)\bigr)y+\mu(\bh)^2}
         {2\sigma^2(\bh)}
  \notag\\
  &=
  -\frac{(y-\mu^*(\bh))^2}{2\sigma^2(\bh)}
  + \frac{(\mu^*(\bh))^2-\mu(\bh)^2}{2\sigma^2(\bh)}
  \notag\\
  &=
  -\frac{(y-\mu^*(\bh))^2}{2\sigma^2(\bh)}
  + \hat\beta_1\mu(\bh)
  + \tfrac{\hat\beta_1^2\sigma^2(\bh)}{2},
  \label{eq:app_complete}
\end{align}
where
\begin{equation}
  \mu^*(\bh)
  :=
  \mu(\bh)+\hat\beta_1\sigma^2(\bh).
  \label{eq:app_mu_star}
\end{equation}
The remaining terms are independent of $y$ and are absorbed into
$\hat Z(\bh)$. Therefore,
\begin{equation}
  \hat p_t(y\mid\bh,\Dtrain)
  =
  \Ncal\!\bigl(y\;\big|\;\mu^*(\bh),\sigma^2(\bh)\bigr).
  \label{eq:app_target_pred}
\end{equation}
Thus, under the estimated log-linear ratio model, exponential tilting shifts
the predictive mean by $\hat\beta_1\sigma^2(\bh)$. The variance is preserved.

The LSA score is the negative log-density of
$\hat p_t(y\mid\bh,\Dtrain)$ up to the additive constant
$\frac{1}{2}\log(2\pi)$:
\begin{multline}
  \slsa(\bh,y)
  =
  -\log \hat p_t(y\mid\bh,\Dtrain)\\
  =
  \frac{(y-\mu^*(\bh))^2}{2\sigma^2(\bh)}
  + \tfrac{1}{2}\log\sigma^2(\bh)
  +\text{const.}
  \label{eq:app_score}
\end{multline}
After dropping the additive constant, this is exactly the form used in
\cref{eq:lsa_gaussian_score}.

\section{Score Inversion}
\label{app:score_derivation}
\paragraph{Exact set inversion.}
For a test input $\bh^*$, the exact sublevel set
$\{y:\slsa(\bh^*,y)\le \hat q\}$ is determined by
\begin{align}
  \frac{(y-\mu^*(\bh^*))^2}{2\sigma^2(\bh^*)}
  + \frac{1}{2}\log\sigma^2(\bh^*)
  &\le \hat q
  \notag\\
  (y-\mu^*(\bh^*))^2
  &\le
  2\sigma^2(\bh^*)\bigl(
    \hat q-\tfrac{\log\sigma^2(\bh^*)}{2}\bigr).
  \label{eq:app_invert}
\end{align}
Hence the exact prediction set is
\begin{equation}
  C_{\mathrm{exact}}(\bh^*)
  =
  {\small\begin{cases}
    \bigl[\mu^*\pm\sigma
      \sqrt{2(\hat q-\tfrac{1}{2}\log\sigma^2)}\bigr],
    & \hat q \ge \tfrac{1}{2}\log\sigma^2(\bh^*),\\[4pt]
    \varnothing,
    & \text{otherwise}.
  \end{cases}}
  \label{eq:app_interval_exact}
\end{equation}

\paragraph{Clipped implementation.}
In our implementation, we use the clipped version
\begin{equation}
  C(\bh^*)
  =
  \Bigl[\mu^*(\bh^*)\pm\sigma(\bh^*)
    \sqrt{2\max\bigl(\hat q-\tfrac{1}{2}\log\sigma^2(\bh^*),0\bigr)}\,\Bigr],
  \label{eq:app_interval}
\end{equation}
which coincides with~\cref{eq:app_interval_exact} whenever the exact
sublevel set is nonempty and avoids numerical issues when the term inside the
square root is negative.

\section{Ablation}

\subsection{Gaussian Mean Weight Estimation}
\label{app:gaussian_mean}

In the main text, density ratios are estimated via logistic regression.
Here we consider an alternative estimator based on Gaussian fitting.
Specifically, we fit separate Gaussian distributions to the source and target
label marginals, obtaining means $\hat{\mu}_s$, $\hat{\mu}_t$ and variances
$\hat{\sigma}_s^2$, $\hat{\sigma}_t^2$. We then construct the estimated
log-density ratio using the fitted means. The source variance is shared:
\begin{align}
  \log \hat r_{\mathrm{GM}}(y)
  &=
  \frac{(y-\hat{\mu}_s)^2-(y-\hat{\mu}_t)^2}
       {2\hat{\sigma}_s^2} \notag\\
  &=
  \frac{(\hat{\mu}_t-\hat{\mu}_s)(2y-\hat{\mu}_s-\hat{\mu}_t)}
       {2\hat{\sigma}_s^2}.
  \label{eq:w_gaussian_mean}
\end{align}
This estimator is linear in $y$, so it preserves the same functional form

\[
  \log \hat r_{\mathrm{GM}}(y)
  =
  \hat\beta_0^{\mathrm{GM}} + \hat\beta_1^{\mathrm{GM}} y
\]
used in the derivation of~\cref{eq:Z_analytic}. The same
closed-form normalizer and the same LSA score construction therefore remain valid after
substituting the corresponding fitted coefficients.

\cref{tab:ablation_gaussian} reports target-domain empirical coverage
and average interval length under this Gaussian-mean weight estimator.
The qualitative trends are consistent with the logistic regression results
in~\cref{tab:main}. The LSA score yields shorter intervals and maintains
comparable empirical coverage under label shift. These results indicate
that the benefits of the LSA score are not tied to a single
weight-estimation procedure. The estimator need only preserve the
log-linear functional form in $y$.

\begin{table*}[t]
\centering
\caption{Coverage and average interval length under
exponential tilt label shift with Gaussian mean weight estimation.
Target coverage is $0.9$.  Results are averaged over $1{,}000$ random
seeds ($\pm$ one standard deviation).
\textbf{Bold} indicates the best value per $\beta$.}
\label{tab:ablation_gaussian}
\begin{tabular}{@{}l ccc ccc@{}}
\toprule
& \multicolumn{3}{c}{Coverage $\uparrow$}
& \multicolumn{3}{c}{Interval Length $\downarrow$} \\
\cmidrule(lr){2-4}\cmidrule(lr){5-7}
$\beta$
  & Res. & Source score & LSA
  & Res. & Source score & LSA \\
\midrule
$0.0$
  & $\mathbf{.9004_{\pm.0094}}$
  & $.9003_{\pm.0094}$
  & $.9002_{\pm.0094}$
  & $3.867_{\pm.099}$
  & $3.841_{\pm.095}$
  & $\mathbf{3.841_{\pm.096}}$ \\
$-0.1$
  & $.8975_{\pm.0126}$
  & $.8973_{\pm.0126}$
  & $\mathbf{.8980_{\pm.0125}}$
  & $4.003_{\pm.120}$
  & $3.974_{\pm.116}$
  & $\mathbf{3.962_{\pm.110}}$ \\
$-0.2$
  & $.8937_{\pm.0153}$
  & $.8937_{\pm.0152}$
  & $\mathbf{.8957_{\pm.0144}}$
  & $4.243_{\pm.166}$
  & $4.207_{\pm.160}$
  & $\mathbf{4.126_{\pm.139}}$ \\
$-0.3$
  & $.8901_{\pm.0201}$
  & $.8899_{\pm.0202}$
  & $\mathbf{.8934_{\pm.0185}}$
  & $4.623_{\pm.253}$
  & $4.577_{\pm.244}$
  & $\mathbf{4.347_{\pm.203}}$ \\
$-0.4$
  & $.8867_{\pm.0271}$
  & $.8863_{\pm.0272}$
  & $\mathbf{.8899_{\pm.0255}}$
  & $5.158_{\pm.419}$
  & $5.091_{\pm.399}$
  & $\mathbf{4.626_{\pm.330}}$ \\
$-0.5$
  & $.8793_{\pm.0394}$
  & $.8792_{\pm.0393}$
  & $\mathbf{.8842_{\pm.0375}}$
  & $5.812_{\pm.659}$
  & $5.740_{\pm.651}$
  & $\mathbf{4.975_{\pm.525}}$ \\
\bottomrule
\end{tabular}
\end{table*}

\subsection{Additional Robustness Study on Lipophilicity}
\label{app:lipophilicity_robustness}

To assess whether the observed effect is specific to AqSolDB, we also
evaluate the same protocol on the Lipophilicity dataset. Despite its smaller
size and different label range, we observe the same qualitative trend. The
three methods are nearly identical at $\beta=0$. Under increasing negative
label shift, the proposed LSA score produces consistently shorter intervals
than both \texttt{Residual} and \texttt{Source score}. At $\beta=-0.5$, the
interval length is reduced by approximately $4.6\%$ relative to Source score
and $5.3\%$ relative to Residual. Empirical coverage remains comparable and
is slightly higher for LSA. This suggests that the effect is not unique to
AqSolDB.

\begin{table*}[t]
\centering
\caption{Results on the Lipophilicity benchmark under exponential-tilt label shift. We report empirical coverage and average interval length (mean $\pm$ standard deviation) over repeated random trials. The proposed LSA score reproduces the same qualitative trend observed on AqSolDB. The three methods are nearly identical at $\beta=0$. Under increasing negative label shift, LSA yields consistently shorter intervals than both residual and source score baselines, with comparable or slightly improved empirical coverage.}
\label{tab:lipophilicity_results}
\begin{tabular}{@{}l ccc ccc@{}}
\toprule
& \multicolumn{3}{c}{Coverage $\uparrow$}
& \multicolumn{3}{c}{Interval Length $\downarrow$} \\
\cmidrule(lr){2-4}\cmidrule(lr){5-7}
$\beta$
  & Res. & Source score & LSA
  & Res. & Source score & LSA \\
\midrule
$0.0$
  & $\mathbf{.9003_{\pm.0151}}$
  & $.9002_{\pm.0152}$
  & $.8997_{\pm.0152}$
  & $2.6333_{\pm.0895}$
  & $2.6170_{\pm.0869}$
  & $\mathbf{2.6163_{\pm.0851}}$ \\
$-0.1$
  & $.8981_{\pm.0197}$
  & $\mathbf{.8983_{\pm.0199}}$
  & $.8976_{\pm.0196}$
  & $2.6742_{\pm.0953}$
  & $2.6580_{\pm.0943}$
  & $\mathbf{2.6476_{\pm.0873}}$ \\
$-0.2$
  & $.8955_{\pm.0205}$
  & $.8957_{\pm.0206}$
  & $\mathbf{.8964_{\pm.0202}}$
  & $2.7229_{\pm.1063}$
  & $2.7071_{\pm.1032}$
  & $\mathbf{2.6799_{\pm.0937}}$ \\
$-0.3$
  & $.8927_{\pm.0216}$
  & $.8925_{\pm.0217}$
  & $\mathbf{.8947_{\pm.0207}}$
  & $2.7847_{\pm.1154}$
  & $2.7684_{\pm.1127}$
  & $\mathbf{2.7155_{\pm.0995}}$ \\
$-0.4$
  & $.8883_{\pm.0248}$
  & $.8885_{\pm.0245}$
  & $\mathbf{.8930_{\pm.0224}}$
  & $2.8556_{\pm.1308}$
  & $2.8379_{\pm.1260}$
  & $\mathbf{2.7499_{\pm.1073}}$ \\
$-0.5$
  & $.8844_{\pm.0271}$
  & $.8847_{\pm.0269}$
  & $\mathbf{.8921_{\pm.0235}}$
  & $2.9404_{\pm.1527}$
  & $2.9215_{\pm.1454}$
  & $\mathbf{2.7860_{\pm.1172}}$ \\
\bottomrule
\end{tabular}
\end{table*}

\section{Encoder Pretraining Details}
\label{app:encoder_details}

All experiments are based on a frozen molecular BART encoder used as the
backbone representation model. The encoder was pretrained on a large-scale
unlabeled molecular corpus assembled from public databases, including
PubChem \citep{kim2023pubchem}, Chembl \citep{zdrazil2024chembl}, Coconut \citep{sorokina2021coconut}, Drugbank \citep{knox2024drugbank}, Enamine \citep{shivanyuk2007enamine}, and ZINC \citep{irwin2012zinc}. After collection and
filtering, the final corpus comprised $212{,}593{,}454$ molecules. The model was pretrained for $5$ epochs and kept fixed throughout all
downstream experiments.

\cref{tab:encoder_details} summarizes the backbone architecture and the main
pretraining hyperparameters.

\begin{table*}[t]
\centering
\caption{Backbone encoder architecture and pretraining setup.}
\label{tab:encoder_details}
\begin{tabular}{@{}lc@{}}
\toprule
Item & Value \\
\midrule
Parameters & $171.16$M \\
Hidden size & $768$ \\
Encoder layers & $8$ \\
Encoder attention heads & $16$ \\
Intermediate size & $3{,}072$ \\
Max position embeddings & $152$ \\
Decoder layers & $8$ \\
Decoder attention heads & $16$ \\
Decoder FFN dimension & $3{,}072$ \\
Vocabulary size & $1{,}016$ \\
Pretraining corpus size & $212{,}593{,}454$ molecules \\
Pretraining epochs & $5$ \\
Data sources & PubChem, ChEMBL, COCONUT, DrugBank, Enamine, ZINC \\
\bottomrule
\end{tabular}
\end{table*}

\end{document}